\documentclass{llncs}

\usepackage{txfonts}

\DeclareMathAlphabet{\mathtt}{OT1}{pcr}{m}{n}
\SetMathAlphabet{\mathtt}{bold}{OT1}{pcr}{b}{n}

\usepackage{graphicx}
\usepackage{amssymb}
\usepackage{algorithm}
\usepackage{algorithmic}
\usepackage{url}

\newcommand{\dataset}{{\cal D}}
\newcommand{\values}{{\cal V}}
\newcommand{\cluster}{{\cal C}}

\newcommand{\cprobs}{{\cal Q}}

\newcommand{\bvec}[1]{\mbox{\boldmath $#1$}}

\newcommand{\fbvec}[1]{\mbox{\footnotesize\boldmath $#1$}}
\newcommand{\vx}{\bvec{x}}
\newcommand{\ve}{\bvec{e}}
\newcommand{\va}{\bvec{a}}

\newcommand{\vxpre}{\vx_\mathrm{prev}}
\newcommand{\vxext}{\vx_\mathrm{ext}}

\newcommand{\fvx}{\fbvec{x}}
\newcommand{\fve}{\fbvec{e}}

\newcommand{\pkx}{p(k\mid\vx)}
\newcommand{\px}{p(\vx)}
\newcommand{\pxk}{p(\vx\mid k)}
\newcommand{\pxnk}{p(\vx\mid\neg k)}
\newcommand{\npkx}{p(k|\vx)}
\newcommand{\npxk}{p(\vx|k)}

\newcommand{\growth}{{\rm GR}}
\newcommand{\pmi}{{\rm PMI}}

\newcommand{\leverage}{{\rm Leverage}}
\newcommand{\tfidf}{\mbox{TF-IDF}}
\newcommand{\gbf}{\mathrm{GBF}}

\newcommand{\ep}{\hat{p}}
\newcommand{\slocal}{s_{\rm local}}
\newcommand{\sglobal}{s_{\rm global}}
\newcommand{\quoted}[1]{\mbox{``$#1$''}}
\newcommand{\algname}[1]{{\sc #1}}

\newcommand{\hq}{\hspace{0.5em}}

\newcommand{\false}{\mathrm{False}}
\newcommand{\is}{\!=\!}

\begin{document}

\title{Verbal Characterization of Probabilistic Clusters using\\Minimal Discriminative Propositions}

\author{
  Yoshitaka Kameya, Satoru Nakamura,
  Tatsuya Iwasaki\thanks{Currently working at NTT Data Corporation} and
  Taisuke Sato
}

\institute{
Graduate School of Information Science and Engineering,
Tokyo Institute of Technology\\
2-12-1 Ookayama, Meguro-ku, Tokyo 152-8552, Japan
}

\maketitle

\begin{abstract}
In a knowledge discovery process, interpretation and evaluation of the mined results are indispensable in practice.  In the case of data clustering, however, it is often difficult to see in what aspect each cluster has been formed.  This paper proposes a method for automatic and objective characterization or ``verbalization'' of the clusters obtained by mixture models, in which we collect conjunctions of propositions (attribute-value pairs) that help us interpret or evaluate the clusters.  The proposed method provides us with a new, in-depth and consistent tool for cluster interpretation/evaluation, and works for various types of datasets including continuous attributes and missing values.  Experimental results with a couple of standard datasets exhibit the utility of the proposed method, and the importance of the feedbacks from the interpretation/evaluation step.
\end{abstract}

\section{Introduction}
\label{sec:intro}

In a knowledge discovery process, interpretation and evaluation of the
mined results are indispensable in practice.
In the case of data clustering~\cite{Jain99}, however, it is often difficult
to see in what aspect each cluster has been formed, only from a list of
the instances in the cluster.  Visualization is a natural way for
understanding things, and particularly in text clustering, Hotho et al.\ applied
formal concept analysis with Hasse diagrams to visualize the similarity
and dissimilarity among the obtained clusters~\cite{Hotho03}.
On the other hand, since there would generally be a physical limitation
or a high implementational cost in visualization, we would rather like to
``verbalize'' the clusters, i.e.\ we associate an intuitive
descriptive label (or a set of such labels) with each cluster.  Additionally
it seems desirable that the labels are chosen objectively and automatically
from the clusters.  So far, there have been only a few labeling methods,
e.g.\ LabelSOM~\cite{Rauber99}, Mei et al.'s automatic labeling for
topic models~\cite{Mei07} and others~\cite{Popescul00,Lamirel08}.
CLIQUE~\cite{Agrawal98} also has a similar
motivation to ours in that it performs hyper-rectangular clustering and
at the same time produces comprehensible descriptions of the obtained clusters.

In this paper, we propose a new labeling method that associates conjunctions
of propositions (attribute-value pairs), called {\em propositional labels},
with the clusters obtained by mixture models.
For example, consider a cluster $\cluster$ which contains several creatures
such as dolphins, mink, platypus and seals.  Then, letting ``milk'' and
``aquatic'' be the boolean attributes of the creatures,
(milk=True $\land$ aquatic=True) would be a suitable
propositional label for the cluster $\cluster$, if none of the
creatures in the other clusters has these properties together.
Finally we easily find that $\cluster$ is a cluster of aquatic mammals.
To find these propositional labels objectively and automatically,
we conduct an Apriori-style breadth-first search for minimal propositional
labels that discriminate the cluster of interest from the others.  Due to
these features, as we will see later, the proposed method can provide us
with a new, in-depth and consistent tool for cluster interpretation/evaluation.
It is also notable that, unlike the previous attempts, the proposed method
is fully applicable to various types of datasets including continuous
attributes and missing values.  Another novel contribution of this paper
is to show empirically the importance of the feedbacks from the
interpretation/evaluation step in achieving a reasonable clustering result.

The rest of this paper is structured as follows.
In Section~\ref{sec:proposal}, we describe the details of the proposed method.
Section~\ref{sec:experiment} then reports the experimental results with
a couple of standard datasets.  Finally, we mention the related work in
Section~\ref{sec:related}, and conclude the paper in
Section~\ref{sec:conclusion}.

\section{Proposed method}
\label{sec:proposal}

\subsection{Preliminaries}
\label{sec:proposal:preliminaries}

Before starting, let us introduce some terminology and notation.
Suppose that we have a dataset $\dataset$ of $N$ instances
which are described by $m$ discrete attributes $A_1,A_2,\ldots,A_m$.
Then, we simply refer to each instance by $\va=(a_1,a_2,\ldots,a_m)$,
where $a_j$ is a value of the $j$-th attribute $A_j$ of the instance.
Also we write $\values(A_j)$ as the set of possible values of $A_j$
(i.e.\ $a_j\in\values(A_j)$, $1\le j\le m$).
We now introduce a propositional label (or a label, for short)
$\quoted{X_1=x_1}\land\quoted{X_2=x_2}\land\cdots\land\quoted{X_n=x_n}$
such that
$\{X_1,X_2,\ldots,X_n\}\subseteq\{A_1,A_2,\ldots,A_m\}$,
$X_i$ and $X_{i'}$ are distinct ($i\neq i'$),
and $x_i\in\values(X_i)$.
In a probabilistic context,
$p(\quoted{X_1=x_1}\land\cdots\land\quoted{X_n=x_n})=
  p(X_1=x_1,\ldots,X_n=x_n)$ holds.
Also, $p(Z=z,\ldots)$ for a random discrete variable $Z$ and its value $z$
is generally abbreviated as $p(z,\ldots)$ if the context is clear.

Furthermore, we add some notational conventions.  First,
without loss of generality, we assume that the attribute values are not
overlapped among attributes
(i.e.\ $\values(A_j)\cap\values(A_{j'})=\emptyset$ for $j\neq j'$).
Then, a propositional label
$\quoted{X_1=x_1}\land\cdots\land\quoted{X_n=x_n}$ is unambiguously simplified
as $\vx=(x_1\land\cdots\land x_n)$ or $\vx=(x_1,\ldots,x_n)$.
Here we have $|\vx|=n$, where $|\vx|$ denotes the number of conjuncts
in $\vx$, and is called the length of $\vx$.
An instance $\va=(a_1,\ldots,a_m)$ is also regarded as
a propositional label $\quoted{A_1=a_1}\land\cdots\land\quoted{A_m=a_m}$.
In this paper, for notational brevity, we use a conjunctive form and a
vector form for propositional labels interchangeably depending on the context.
Besides, to simplify the algorithm descriptions presented later,
in a propositional label
$\quoted{X_1=x_1}\land\cdots\land\quoted{X_n=x_n}$,
we will always enumerate $X_1,X_2,\ldots$ so that the order
of enumeration preserves the original one $A_1,A_2,\ldots$,
i.e.\ for $j_1$, $j_2$, \ldots $j_n$ such that $X_i$ corresponds to $A_{j_i}$
($1\le i\le n$, $1\le j_i\le m$), $j_i<j_{i'}$ holds when $i<i'$.

Here consider a propositional label $\vx=(x_1,\ldots,x_n)$.
Then, a label $\vx'=(x'_1,\ldots,x'_{n'})$ is called a {\em subconjunction}
of $\vx$ if $\{x'_1,\ldots,x'_{n'}\}\subseteq\{x_1,\ldots,x_n\}$,
and we denote this by $\vx'\subseteq\vx$.  If $\vx'\subseteq\vx$
but $\vx'\neq\vx$, we write $\vx'\subset\vx$.
For an instance $\va$ and a propositional label $\vx$,
we say ``$\va$ satisfies $\vx$'' if $\vx\subseteq\va$.
For a boolean attribute $A_j$, we may abbreviate $\quoted{A_j\is\mathrm{True}}$
and $\quoted{A_j\is\mathrm{False}}$ as
$\quoted{A_j\is\mathrm{T}}$ and $\quoted{A_j\is\mathrm{F}}$, respectively.

\subsection{Overview}
\label{sec:proposal:overview}

In this paper, we consider probabilistic clustering based on a
simple mixture model called a naive Bayes model.  A naive Bayes model
has a latent class variable $C$ taking on the identifiers $\{1,2,\ldots,K\}$
of $K$ clusters, and represents a simple joint distribution:
$p(C\is k,A_1\is a_1,\ldots,A_m\is a_m)=
    p(C\is k)\prod_{j=1}^m P(A_j\is a_j\mid C\is k)$,
or equivalently $p(k,\va)=p(k)\prod_j p(a_j\mid k)$.
Here the probabilities $p(k)$ and $p(a_j\mid k)$ are treated
as the model parameters.
Given a dataset $\dataset$ of instances and the number $K$ of clusters, we do:

\begin{enumerate}
\item
  Estimate the parameters in a model $p(k,\va)$ from $\dataset$.
\item
  Assign the most probable class
  $k^\ast(\va)={\rm argmax}_{1\le k\le K}$ $p(k\mid\va)$
  to each instance $\va$ based on the estimated parameters.
  The $k$-th cluster $\cluster_k$ is then formed as
  a set of instances $\va$ such that $k^\ast(\va)=k$.
\item
  Find propositional labels $\vx$ that characterize well
  each cluster $\cluster_k$.
\end{enumerate}

\noindent
In the first two steps, we perform clustering, and the third step is called
{\em labeling}.  As is well-known, the first step is realized by the EM
(expectation-maximization) algorithm~\cite{Dempster77}.\footnote{
As discussed in Section~\ref{sec:related}, 
we can also use the $K$-means algorithm for clustering.
}
From the second step, clustering can be casted as an unsupervised
classification task, and we call $p(k\mid\va)$ the
{\em (class) membership probability} of an instance $\va$.
In the last step, it is unspecified what are the propositional labels
that characterize the clusters, and how to obtain them.
The next two sections, Sections~\ref{sec:proposal:cpl} and
\ref{sec:proposal:search}, address these issues, respectively.

\subsection{Characteristic propositional labels}
\label{sec:proposal:cpl}

\subsubsection{Relevance scores:}
\label{sec:proposal:cpl:relv}

To choose suitable propositional labels
$\vx=(x_1\land\cdots\land x_n)$ or
$\vx=(x_1,\ldots,x_n)$, of a cluster $\cluster_k$
objectively and automatically, we introduce a scoring function
that measures how relevant $\vx$ and $\cluster_k$ are.
Previously, several {\em relevance scores} have been proposed in
various statistical/data-mining tasks.
The followings are an adaptation of those relevance scores to our
labeling problem:
\begin{itemize}
\item
  Growth rate:\ $\growth_k(\vx)=\pxk/\pxnk$, where
  $\neg k$ indicates that the instance under consideration
  belongs to a class other
  than $k$.  This score is mainly used in emerging pattern
  mining~\cite{Dong99} and explicitly states that the instances satisfying
  $\vx$ are likely to occur in the cluster $\cluster_k$ and unlikely to
  occur in the clusters other than $\cluster_k$.
  $\growth_k(\vx)$ ranges from 0 (when $\pxk=0$) to $\infty$
  (when $\pxk>0$ and $\pxnk=0$).
\par\vspace{5pt}\par
\item
  Membership probabilities:\ $\pkx$.
  PRIM, a rule-based method for bump hunting, tries to find $\vx$
  such that $\pkx\ge r$,
  where $r$ is some threshold, under a separate-and-conquer
  strategy~\cite{Friedman99}.
  It is crucial to see that, for a fixed $k$, 
  $\pkx=p(k)\pxk/\px\propto\pxk/\px$ holds.
  In class association rule (CAR) mining~\cite{Liu98},
  $\pkx$ is called the {\em confidence} of a rule $\vx\Rightarrow\cluster_k$.
\par\vspace{5pt}\par
\item
  Pointwise mutual information:\ $\pmi_k(\vx)=\log p(k,\vx)-\log\{p(k)\px\}$.
  PMI has been used in text analysis~\cite{Church89}.
  This score is rewritten as $\log\pxk-\log\px$, which is
  adopted by a well-known probabilistic clustering tool
  AutoClass~\cite{Cheeseman95} for post-analysis
  (named ``attribute influence values''),
  in a limited case with $|\vx|=1$.
  The non-logarithmic version $p(k,\vx)/(p(k)\px)$ is called the
  {\em lift} of
  a class association rule $\vx\Rightarrow\cluster_k$~\cite{Geng06}.
\par\vspace{5pt}\par
\item
  Leverage:\ $\leverage_k(\vx)=p(k,\vx)-p(k)p(\vx)$.
  This score is often used for finding interesting
  association rules~\cite{Webb03}.
  $\leverage_k(\vx)$ is equivalent to the {\em weighted relative accuracy}
  (WRAcc), a score used in subgroup discovery, and can be rewritten as
  $\px(\pkx-p(k))$ or
  $p(k)p(\neg k)(\pxk - p(\vx\mid\neg k))$~\cite{KraljNovak09}.
  A related score $|\pxk - p(\vx\mid\neg k)|$, often called
  {\em support difference}, is used in contrast set mining~\cite{Bay01}.
\par\vspace{5pt}\par
\item
  TF-IDF:\ $\tfidf_k(\vx)=\pxk\log\{1/\px\}$.
  This is a popular measure in information retrieval~\cite{Manning08},
  and is a product of term frequency (TF) and inverse document
  frequency (IDF).
  TF of a term $t$ in a document $d$ is the relative
  frequency of $t$ occurring in $d$, and IDF of $t$ is the logarithm
  of the inverse of the relative frequency that a document containing $t$
  occurs in the whole document set.
  Then, assuming that a term occurs at most once
  in a document, the TF-IDF of a term $t$ in a document $d$
  is given as $p(t\mid d)\log\{1/p(t)\}$.  Since TF-IDF is known
  to give a reasonably high score to $t$
  that characterizes $d$,
  $\mbox{TF-IDF}_k(\vx)$ above
  can be used by analogy where $t$ corresponds to $\vx$, and $d$
  corresponds to $k$.
\par\vspace{5pt}\par
\item
  Precision/Recall: Precision and recall are also popular measures in
  information retrieval.  In our context, $\pkx$ and $\pxk$
  can be regarded as precision and recall of label $\vx$ for the $k$-th
  cluster~\cite{Geng06}. Also in COBWEB~\cite{Fisher87}, a well-known
  conceptual clustering method, $\pkx$ and $\pxk$ are respectively used
  as metrics for inter-class dissimilarity and intra-class similarity.
  To balance the opposite behavior of precision and recall,
  in information retrieval, we often use
  thier harmonic mean $2\pkx\pxk/(\pkx + \pxk)$ and call it the {\em F-score}.
  Lamirel et al.\ proposed the use of the F-score for automatic labeling
  of clustering results~\cite{Lamirel08}.
  Similarly, the product of precision and recall $\pkx\pxk$,
  which substantially works as the geometric mean of $\pkx$ and $\pxk$,
  is used by Popescul and Ungar~\cite{Popescul00}.
\end{itemize}

\noindent
Other relevance scores are discussed in comprehensive surveys by
Kralj Novak et al.~\cite{KraljNovak09} and by Geng et al.~\cite{Geng06}.
It is easy to show that
$p(k\mid\vx_1)\le p(k\mid\vx_2)$ iff
$\growth_k(\vx_1)\le\growth_k(\vx_2)$,\footnote{
$\growth_k(\vx)=(p(k)/p(\neg k))^{-1}(\pkx/p(\neg k\mid\vx))\propto\pkx/(1-\pkx)$.
}
and
$p(k\mid\vx_1)\le p(k\mid\vx_2)$ iff
$\pmi_k(\vx_1)\le\pmi_k(\vx_2)$.
Consequently, for a particular cluster $\cluster_k$,
the first three scores give the same ranking over the propositional labels.
Hereafter we call $\pxk$ the {\em local support}, and $\px$
the {\em global support}.  The relevance scores above commonly
rely on the local support with a penalty regarding the global support.
This contrastive use of the global support and the local support is also
found in the category utility adopted in COBWEB~\cite{Fisher87}.

In this paper, we choose $\pkx$ as the relevance score 
for two reasons on intuitiveness for the end users.
First, we can of course interpret $\pkx$ as discriminative probabilities,
by which we classify an instance satisfying $\vx$.
As mentioned in Section~\ref{sec:proposal:overview}, clustering is
performed based on the membership probabilities $p(k\mid\va)$,
which are a special case of $p(k\mid\vx)$.
The second reason is more practical:\ $\pkx$ is inherently
normalized (i.e.\ $0\le \pkx\le 1$).  From this nature, we can use
a threshold $r$, which just ranges over $(0,1]$ and is commonly
applied to all clusters, to filter out $\vx$ such that $\pkx<r$.

\subsubsection{Minimality:}
\label{sec:proposal:cpl:minimal}

Let us consider two propositional labels $\vx_1$ and $\vx_2$ that
fulfill some requirement (e.g.\ $p(k\mid\vx_1)\ge r$ and $p(k\mid\vx_2)\ge r$
for some threshold $r$), and also suppose that $\vx_1\subseteq\vx_2$ holds.
In such a case, we favor $\vx_1$ over $\vx_2$, because
the longer one may have some redundant information which hinders us
from understanding the cluster.
In other words, we would like to have only {\em minimal} labels.
In the literature on emerging pattern mining, such minimal
patterns are called essential emerging patterns~\cite{Fan03},
and Ji et al.\ proposed an efficient mining algorithm named
{\em ConSGapMiner} for minimal distinguishing sequences~\cite{Ji07}.

\subsubsection{Model-based computation of relevance scores:}
\label{sec:proposal:cpl:model}

We have introduced several relevance scores which are based on
probabilities.  In most of the previous work,
these probabilities are directly estimated from
a given dataset $\dataset$ of instances.  For example,
membership probabilities are estimated as
$\ep(k\mid\vx)=
    |\{\va\in\cluster_k\mid\vx\subseteq\va\}|\;/\;
    |\{\va\in\dataset\mid\vx\subseteq\va\}|$.
In our method, on the other hand, relevance scores are computed
from the model parameters via the joint distribution
(Section~\ref{sec:proposal:overview}).
This model-based approach has a couple of advantages.
First, as seen later, we can efficiently compute the scores, exploiting
the conditional independence in the model, without scanning the whole
dataset $\dataset$.
In many cases,
the space for the model parameters is much smaller than the dataset.
The second advantage is that the model parameters are
well-abstracted data as long as the model fits to $\dataset$,
and there would be less chance to be affected by noise.
Finally, there is a positive side-effect that we need not care about
missing values in $\dataset$ since we only use the parameters
estimated by the EM algorithm.

\subsubsection{Selecting characteristic propositional labels:}
\label{sec:proposal:score:task}

Now based on the discussions above, we define
{\em characteristic propositional labels}, which characterize well
the obtained clusters.  A propositional label
$\vx$ of the cluster $\cluster_k$ is characteristic iff:

\begin{enumerate}
\item $\pkx\ge r$,\label{enum:clabel:relv}
\item $\px\ge\sglobal$,\label{enum:clabel:gsupport}
\item $\pxk\ge\slocal$, and\label{enum:clabel:lsupport}
\item There is no $\vx'\subset\vx$ that satisfies 1$\sim$3 above,
  \label{enum:clabel:min}
\end{enumerate}
where $r$, $\sglobal$ and $\slocal$
are user-specified thresholds, and the probabilities
$\pkx$, $\px$ and $\pxk$ are computed via the joint distribution.
Conditions~\ref{enum:clabel:relv}$\sim$\ref{enum:clabel:min}
are called
the {\em relevance condition},
the {\em global support condition},
the {\em local support condition}, and
the {\em minimality condition},
respectively.

While most of the existing CAR mining algorithms run based on
the guide from the threshold for $\pxk$,
we treat the first and the fourth conditions as the primary filters.
The remaining conditions are introduced to remedy the problem
that we often obtain unintuitive characteristic labels
with very low global/local support,
and also to reduce the burden in the exhaustive search for
characteristic labels, which will be described in the next section.
So currently we do not consider to put a tight restriction on
global/local support
(e.g.\ $\slocal=1/(|\dataset|/K)=K/|\dataset|$, which implies that
each of equally-sized clusters should contain at least one instance).

\subsection{Exhaustive search for characteristic propositional labels}
\label{sec:proposal:search}

All possible propositional labels form a version space~\cite{Mitchell97},
and on this structure, we conduct an Apriori-style breadth-first
search for the entire set of characteristic
labels for each cluster.  There are two major styles for such an
exhaustive search:\ depth-first
and breadth-first.  We take a breadth-first style
because, as seen later, it is easier to check
the minimality of characteristic labels in a breadth-first style,\footnote{
{\em ConSGapMiner} mentioned above works in a depth-first fashion,
and needs to introduce an extra data structure (a prefix tree) to reduce
the time for the post-check on minimality.
}
and because we do not necessarily need very long characteristic labels
that are difficult to read.

The \algname{Find} procedure (Algorithm~\ref{alg:Find}) is the main routine
of the search algorithm for characteristic labels, which
calls the \algname{GenCandidate} function (Algorithm~\ref{alg:GenCandidate}).
The basic flow is similar to Apriori (\algname{GenCandidate}
is our version of the {\tt apriori-gen} function in \cite{Agrawal94}),
but is different in that we make probability computation while generating
candidates.  In addition, since this probability computation requires
normalization for each membership probability $\pkx$, the most part of
the algorithm should work in parallel for clusters.
It is also crucial to note that the global/local support of $\vx$
($\px$ and $\pxk$) are anti-monotonic w.r.t.\ the inclusion relation
(i.e.\ $\pxk\ge p(\vx'\mid k)$ if $\vx\subseteq\vx'$), but in general
our relevance score is not.
Instead, like {\em ConSGapMiner}, we make pruning based on
the minimality of characteristic labels.

In the \algname{Find} procedure, for each $\cluster_k$,
$S_n[k]$ indicates a set of propositional labels
of length $n$ that satisfy the global/local support condition,
and $R_n[k]$ indicates a set of labels in $S_n[k]$ that additionally
satisfy the relevance condition.  $R_n[k]$ are the characteristic labels
of length $n$ which we wish to have, and we do not extend the labels in 
$R_n[k]$.  $W_n[k]=S_n[k]\setminus R_n[k]$ are therefore
the labels to be worked on next.

The candidate labels of length $(n+1)$ are generated
from the \algname{GenCandidate} function, in which the labels of length $n$
in $W_n[k]$ are combined effectively.  In Line~\ref{alg:GenCandidate:prune}
of \algname{GenCandidate}, like the ``prune'' step of Apriori,
``$\mbox{\sc SubConj}(\vxext)\subseteq W_n[k]$''
filters out the over-generated candidate labels using
anti-monotonicity of global/local support and minimality at the same time.
$\mbox{\algname{SubConj}}(\vx)$ is a function that returns a set
of $\vx$'s subconjunctions of
length $|\vx|-1$,\footnote{
More specifically, for $\vx=(x_1,\ldots,x_{n-1},x_n)$,
$\mbox{\algname{SubConj}}(\vx)=$\linebreak[3]
$\{(x_2,\linebreak[3]x_3,\linebreak[3]\ldots,\linebreak[3]x_{n-1},\linebreak[3]x_n)$,\linebreak[3]
$(x_1,\linebreak[3]x_3,\linebreak[3]\ldots,\linebreak[3]x_{n-1},\linebreak[3]x_n)$,\linebreak[3]
\ldots,
$(x_1,\linebreak[3]x_2,\linebreak[3]\ldots,\linebreak[3]x_{n-2},\linebreak[3]x_n)$,
$(x_1,\linebreak[3]x_2,\linebreak[3]\ldots,\linebreak[3]x_{n-2},\linebreak[3]x_{n-1})\}$.
}
and using the property that $W_n[k]=S_n[k]\setminus R_n[k]$, the filtering
condition requires that each of the immediate subconjunctions of $\vxext$
should be in $S_n[k]$ (due to anti-monotonicity),
but should not be in $R_n[k]$ (due to minimality).
This way of filtering, together with the breadth-first strategy,
enables us to perform effective pruning by only checking the labels
in $W_n[k]$.\footnote{
$W_n[k]$ is constructed from the labels in $W_{n-1}[k]$, and hence
is guaranteed not to include any labels $\vx$ such that $\vx'\subseteq \vx$,
$\vx'\in R_{n'}[k]$ and $1\le n'\le n$.
}
Then, for each candidate label that has passed the filter,
we compute the probabilities $\pxk$, $\px$ and $\pkx$
(Lines~\ref{alg:GenCandidate:prob:begin}--\ref{alg:GenCandidate:prob:end}).
The point here is that we take the union of $C_{n+1}[k]$'s in advance
(Line~\ref{alg:GenCandidate:union}) to avoid a redundant computation,
and reuse the previously computed values for $p(\vxpre\mid k)$,
exploiting the conditional independence in the naive Bayes model.

\begin{algorithm}[t]
\caption{\algname{Find}}
\label{alg:Find}
\begin{small}
\begin{algorithmic}[1]
\FORALL{$k=1,2,\ldots,K$}
\STATE
  $S_1[k]:=
    \{a_j\mid 1\le j\le m,\;a_j\in\values(A_j), p(a_j)\ge\sglobal,\;p(a_j\mid k)\ge\slocal\}$
\label{alg:Find:s1}
\STATE $R_1[k]:=\{a_j\in S_1[k]\mid p(k\mid a_j)\ge r\}$
\label{alg:Find:r1}
\STATE $W_1[k]:=S_1[k]\setminus R_1[k]$
\ENDFOR
\STATE
\STATE $n:=1$
\WHILE{$\exists k: W_n[k]\neq\emptyset$}
\STATE $\langle C_{n+1}[1],\ldots,C_{n+1}[K]\rangle:=\mbox{\sc GenCandidate}(W_n[1],\ldots,W_n[K])$
\FORALL{$k=1,2,\ldots,K$ such that $C_{n+1}[k]\neq\emptyset$}
\STATE $S_{n+1}[k]:=\{\vx\in C_{n+1}[k]\mid\px\ge\sglobal,\;\pxk\ge\slocal\}$
\label{alg:Find:s2}
\STATE $R_{n+1}[k]:=\{\vx\in S_{n+1}[k]\mid\pkx\ge r\}$
\label{alg:Find:r2}
\STATE $W_{n+1}[k]:=S_{n+1}[k]\setminus R_{n+1}[k]$
\ENDFOR
\STATE $n:=n+1$
\ENDWHILE
\STATE
\RETURN $\langle\bigcup_n R_n[1],\;\ldots,\;\bigcup_n R_n[K]\rangle$
\end{algorithmic}
\end{small}
\end{algorithm}

To speed-up further the search algorithm in the case with many attributes,
in the \algname{Find} procedure,
we optionally introduce a greedy pruning,
similarly to a commercial data-mining tool named Magnum Opus~\cite{Webb03}.
To be more concrete, we delete $a_j$ such that $p(k\mid a_j)<p(k)$
from $S_1[k]$ after Line~\ref{alg:Find:s1}.
In addition, $\vx=(x_1,\ldots,x_n,x_{n+1})\in S_{n+1}[k]$
such that $p(k\mid\vx)<p(k\mid\vx')$,
where $\vx'=(x_1,\linebreak[3]\ldots,\linebreak[3]x_n)$, are
considered as unpromising, and deleted from $S_{n+1}[k]$
after Line~\ref{alg:Find:s2}.  This greedy pruning is unsafe, i.e.\ we may
miss some characteristic labels actually satisfying the conditions
in Section~\ref{sec:proposal:score:task}, but it would bring high
efficiency in many practical cases.

\begin{algorithm}[t]
\caption{\sc GenCandidate($W_n[1],\ldots,W_n[K]$)}
\label{alg:GenCandidate}
\begin{small}
\begin{algorithmic}[1]
\FORALL{$k=1,2,\ldots,K$}
\STATE $C_{n+1}[k]:=\emptyset$
\FORALL{
  $\vx=(x_1,\ldots,x_{n-1},x_n)\in W_n[k]$ and $\vx'=(x_1,\ldots,x_{n-1},x'_n)\in W_n[k]$
  such that $\forall j$:\ $x_n,x'_n\not\in\values(A_j)$
}
\STATE $\vxext:=(x_1,\ldots,x_{n-1},x_n,x'_n)$
\IF{
  $\mbox{\sc SubConj}(\vxext)\subseteq W_n[k]$
}
\label{alg:GenCandidate:prune}
\STATE $C_{n+1}[k]:=C_{n+1}[k]\cup\{\vxext\}$
\ENDIF
\ENDFOR
\ENDFOR
\STATE
\STATE $D_{n+1}:=\bigcup_{k=1}^K C_{n+1}[k]$
\label{alg:GenCandidate:union}
\label{alg:GenCandidate:prob:begin}
\FORALL{$\vx=(x_1,\ldots,x_{n-1},x_n,x_{n+1})\in D_{n+1}$}
\STATE $\vxpre:=(x_1,\ldots,x_n)$
\STATE $p(\vx\mid k):=p(\vxpre\mid k)p(x_{n+1}\mid k)$
\hq for $k=1,\ldots,K$
\STATE $p(\vx):=\sum_{k=1}^K p(k)p(\vx\mid k)$
\ENDFOR
\STATE
\STATE $\displaystyle p(k\mid\vx):=p(k)p(\vx\mid k)/p(\vx)$
    for $k=1,\ldots,K$ and $\vx\in C_{n+1}[k]$
\label{alg:GenCandidate:prob:end}
\STATE
\RETURN $\langle C_{n+1}[1],\ldots,C_{n+1}[K]\rangle$
\end{algorithmic}
\end{small}
\end{algorithm}

\subsection{Handling continuous attributes}
\label{sec:proposal:cont}

Until now, we have assumed that all attributes are discrete.
To handle continuous attributes and discrete attributes consistently
in terms of membership probabilities,
we also ``propositionalize'' each continuous attribute.
To be more specific, as is often done in mixture modeling, we
consider that each continuous
attribute follows a univariate Gaussian distribution, in which
two types of parameters, the mean $\mu_{j,k}$ and the variance $\sigma^2_{j,k}$,
are introduced for the $j$-th continuous attribute $A_j$ and the $k$-th
cluster $\cluster_k$.
These parameters are also estimated by the EM algorithm. We further
assume that we are given a set $\cprobs=\{q_1,q_2,\ldots,q_{|\cprobs|}\}$
of different probabilities, where $0<q_h<1$ for $1\le h\le |\cprobs|$,
and the indices are given so that $q_h<q_{h'}$ if $h<h'$.
For instance, we may have $\cprobs=\{0.1,0.2,\ldots,0.9\}$.
Then, using a cumulative distribution function $F_{j,k}$ with the mean
$\mu_{j,k}$ and the variance $\sigma^2_{j,k}$ for each $A_j$ and $\cluster_k$,
we introduce $\quoted{\alpha_h^{(j,k)}< A_j\le \beta_h^{(j,k)}}$ as a conjunct
in a propositional label, where $\alpha_h^{(j,k)}=\mu_{j,k}-d_h^{(j,k)}$
and $\beta_h^{(j,k)}=\mu_{j,k}+d_h^{(j,k)}$ such that
$F_{j,k}(\beta_h^{(j,k)})-F_{j,k}(\alpha_h^{(j,k)})=q_h$.
It can be seen here that $\alpha_h^{(j,k)}$ and $\beta_h^{(j,k)}$ are symmetric
w.r.t.\ the mean $\mu_{j,k}$.  Hereafter we omit the superscript
${(j,k)}$ unless they are needed.

Now consider $\vx=(\vx_0\land\quoted{\alpha_h< X_n\le \beta_h})$
and $\vx'=(\vx_0\land\quoted{\alpha_{h'}< X_n\le \beta_{h'}})$
where $h<h'$.  Then, we define $\vx'\subset\vx$ and we have
$p(\vx)<p(\vx')$.  With this new inclusion relation, in the search
algorithm, an additional minimality check is made
for the last conjunct corresponding to a continuous attribute,
just after $R_n[k]$ being computed.\footnote{
To be specific, after Lines~\ref{alg:Find:r1} and \ref{alg:Find:r2}
in the \algname{Find} procedure, non-minimal labels
are deleted from both $R_n[k]$ and $S_n[k]$.
}
One may see that $\alpha_h$'s and $\beta_h$'s above are model-based
quantile values,\footnote{
For instance, if $q_h$ is given as 0.9, $\alpha_h$ and $\beta_h$ respectively
correspond to the 5\%-tile value and the 95\%-tile value under the
Gaussian distribution.
}
and choosing an appropriate $\quoted{\alpha_h< A_j\le \beta_h}$
leads to an automatic adjustment of $(\alpha_h,\beta_h)$, which
resembles the `peeling' operation in PRIM, a rule-based bump hunting
method~\cite{Friedman99}.

\section{Experiments}
\label{sec:experiment}

In the experiments, we used four datasets:\ the zoo dataset, 
the iris dataset, the 20 newsgroup dataset and the flags dataset.\footnote{
The zoo dataset, the iris dataset and the flags dataset are available from
the the UCI ML Repository
({\tt http://\linebreak[3]archive.\linebreak[3]ics.\linebreak[3]uci.\linebreak[3]edu/\linebreak[3]ml/}),
and the 20 newsgroup dataset is
available from the UCI KDD Archive
({\tt http://\linebreak[3]kdd.\linebreak[3]ics.\linebreak[3]uci.\linebreak[3]edu/}).
}
For the first three datasets, we gave the correct number $K$ of clusters
to the clustering algorithm, considering ideal situations.
We then compare the obtained characteristic labels and
the original (human-annotated) classes.  On the other hand,
since the flags dataset does not contain the class information,
we explore a plausible number of clusters by characteristic labels
together with a Bayesian score for model selection.
For simplicity, throughout the experiments, we set a small value
($1/|\dataset|$) to the threshold $\sglobal$ for the global support $\px$,
so that the influence from $\sglobal$ is negligible.
In addition, we tried 1,000 re-initializations
in the EM algorithm not to get trapped into unwanted local optima.

\subsection{Zoo dataset}
\label{sec:experiment:zoo}

The zoo dataset describes the classification of 101 species of creatures
with 17 attributes.  The species are originally categorized into seven classes.
Table~\ref{tab:zoo:1} (top-left) shows the confusion matrix of
the clustering result.  We can see from this matrix that the creatures in
the class ``mammals'' are split into two clusters $\cluster_1$ and $\cluster_2$,
whereas the creatures in ``reptiles'' and ``amphibians'' are merged
into cluster $\cluster_5$.
\begin{table}[t]
\caption{
The confusion matrix,
and the characteristic labels for the clusters $\cluster_1$,
$\cluster_2$ and $\cluster_3$ in the zoo dataset.
}
\label{tab:zoo:1}
\begin{center}

\begin{tabular}{cc}
{
\tabcolsep=2.5pt
\begin{tabular}[t]{|r|ccccccc|}
\hline
&\multicolumn{7}{|c|}{clusters}\\
original classes&$\cluster_1$&$\cluster_2$&$\cluster_3$&$\cluster_4$&$\cluster_5$&$\cluster_6$&$\cluster_7$\\
\hline
mammals&35&6&0&0&0&0&0\\
birds&0&0&20&0&0&0&0\\
fishes&0&0&0&13&0&0&0\\
amphibians&0&0&0&0&4&0&0\\
reptiles&0&0&0&0&5&0&0\\
insects&0&0&0&0&0&8&0\\
others&0&0&0&0&1&2&7\\
\hline
\end{tabular}
}
&
{\footnotesize
\renewcommand{\arraystretch}{0.9}
\tabcolsep=2.0pt
\begin{tabular}[t]{|lcl|r|r|r|}
\hline
\multicolumn{3}{|c|}{labels for $\cluster_1$}&$\npkx$&$\npxk$\\
\hline
milk=T&$\wedge$&aquatic=F&1.000&1.000\\
eggs=F&$\wedge$&aquatic=F&0.972&1.000\\
milk=T&$\wedge$&fins=F&0.945&1.000\\
hair=T&$\wedge$&toothed=T&0.913&1.000\\
hair=T&$\wedge$&eggs=F&0.913&1.000\\
eggs=F&$\wedge$&fins=F&0.905&1.000\\
hair=T&$\wedge$&tail=T&0.900& 0.857\\
hair=T&$\wedge$&legs=4&0.956&0.828\\
milk=T&$\wedge$&legs=4&0.935&0.828\\
eggs=F&$\wedge$&legs=4&0.910&0.828\\
&\multicolumn{1}{c}{:}&&\multicolumn{1}{|c|}{:}&\multicolumn{1}{c|}{:}\\
\hline
\end{tabular}
}
\\
{\footnotesize
\renewcommand{\arraystretch}{0.9}
\tabcolsep=2.0pt
\begin{tabular}[t]{|lcl|r|r|r|}
\hline
\multicolumn{3}{|c|}{labels for $\cluster_2$}&$\npkx$&$\npxk$\\
\hline
milk=T&$\wedge$&aquatic=T&1.000&1.000\\
breathes=T&$\wedge$&fins=T&1.000&0.666\\
milk=T&$\wedge$&fins=T&1.000&0.666\\
hairs=T&$\wedge$&aquatic=T&1.000&0.666\\
eggs=F&$\wedge$&fins=T&1.000&0.555\\
milk=T&$\wedge$&legs=0&1.000&0.500\\
hairs=T&$\wedge$&fins=T&1.000&0.444\\
hairs=F&$\wedge$&milk=T&1.000&0.333\\
fins=T&$\wedge$&legs=4&1.000&0.222\\
&\multicolumn{1}{c}{:}&&\multicolumn{1}{|c|}{:}&\multicolumn{1}{c|}{:}\\
\hline
\end{tabular}
}
&
{\footnotesize
\renewcommand{\arraystretch}{0.9}
\tabcolsep=2.0pt
\begin{tabular}[t]{|lcl|r|r|r|}
\hline
\multicolumn{3}{|c|}{labels for $\cluster_3$}&$\npkx$&$\npxk$\\
\hline
feathers=T&&&1.000&1.000\\
milk=F&$\wedge$&legs=2&1.000&1.000\\
toothed=F&$\wedge$&legs=2&0.991&1.000\\
eggs=T&$\wedge$&legs=2&0.991&1.000\\
hairs=F&$\wedge$&legs=2&0.983&1.000\\
airborne=T&$\wedge$&legs=2&0.979&0.800\\
airborne=T&$\wedge$&tail=T&0.903&0.800\\
legs=2&$\wedge$&catsize=F&0.900&0.700\\
airborne=T&$\wedge$&aquatic=T&1.000&0.240\\
&\multicolumn{1}{c}{:}&&\multicolumn{1}{|c|}{:}&\multicolumn{1}{c|}{:}\\
\hline
\end{tabular}
}
\end{tabular}
\end{center}
\end{table}
Besides, the remaining tables in
Table~\ref{tab:zoo:1} show the obtained characteristic labels
for $\cluster_1$, $\cluster_2$ and $\cluster_3$,
where $\cluster_3$ corresponds to the original class ``birds.''
The labels in the tables are ordered firstly according to the length of
$\vx$ (i.e.\ syntactic generality), secondly according to the magnitude
of $\pxk$ (i.e.\ statistical generality), and thirdly
according to the magnitude of $\pkx$.\footnote{
We also observed that intuitive labels tend to be highly
ranked according to the harmonic mean of $\pkx$ and $\pxk$.
}
We used $r=0.9$ and $\slocal=K/|\dataset|$
as the thresholds for $\pkx$ and $\pxk$, respectively, where
$\dataset$ is the dataset and $K=3$ is the number of clusters.

Since the original classes are unknown in real situations,
we interpret the clusters $\cluster_1$, $\cluster_2$ and
$\cluster_3$, only from the obtained characteristic labels.
For example, all creatures in $\cluster_3$ have feathers, so we can guess
that $\cluster_3$ corresponds to birds. Also there are several plausible labels
for $\cluster_3$ which support our guess.
Interestingly, on the other hand,
the obtained labels indicate that the (wrongly) split classes $\cluster_1$
and $\cluster_2$ correspond to terrestrial and aquatic mammals, respectively.
So one may conclude that these split clusters are still meaningful.
In the past, to evaluate the quality of the obtained clusters, there
has been no way but to numerically check the closeness between the
obtained clusters
and the human-annotated classes, using some matching criteria,
such as purity, normalized
mutual information and the (adjusted) Rand index~\cite{Manning08,Meila07}.
Contrastingly, as seen above, the characteristic labels provide
us with a new and in-depth way for cluster evaluation.
Similar interpretations are possible for the other clusters,
whose characteristic labels are shown in Table~\ref{tab:zoo:2}.

\begin{table}[t]
\caption{
The obtained characteristic labels for clusters $\cluster_4$,
$\cluster_5$, $\cluster_6$ and $\cluster_7$.
}
\label{tab:zoo:2}
\centerline{
\tabcolsep=1.0pt
\begin{tabular}{cc}
{
\footnotesize
\renewcommand{\arraystretch}{0.9}
\tabcolsep=1.0pt
\begin{tabular}{|lcl|r|r|}
\hline
\multicolumn{3}{|c|}{labels for $\cluster_4$}&$\npkx$&$\npxk$\\
\hline
milk=F&$\land$&fin=T&1.000&1.000\\
breathes=F&$\land$&tail=T&0.948&1.000\\
eggs=T&$\land$&fin=T&0.951&1.000\\
toothed=T&$\land$&breathes=F&0.941&1.000\\
backbone=T& $\land$&breathes=F&0.935&1.000\\
breathes=F&$\land$&fin=T&1.000&1.000\\
hair=F& $\land$&fin=T&0.906&1.000\\
fin=T&$\land$&catsize=F&1.000&0.692\\
\multicolumn{3}{|c}{:}&\multicolumn{1}{|c|}{:}&\multicolumn{1}{c|}{:}\\
\hline
\end{tabular}
}
&
{
\footnotesize
\renewcommand{\arraystretch}{0.9}
\tabcolsep=1.0pt
\begin{tabular}{|lcl|r|r|}
\hline
\multicolumn{3}{|c|}{labels for $\cluster_7$}&$\npkx$&$\npxk$\\
\hline
legs=5&&&1.000&0.142\\
backbone=F&$\land$&breathes=F&0.985&1.000\\
toothed=F&$\land$&breathes=F&0.972&1.000\\
breathes=F&$\land$&tail=F&0.958&1.000\\
aquatic=T&$\land$&backbone=F&0.922&0.857\\
breathes=F&$\land$&legs=6&1.000&0.285\\
aquatic=T&$\land$&legs=6&1.000&0.245\\
backbone=F&$\land$&legs=8&0.908&0.142\\
backbone=F&$\land$&catsize=T&0.908&0.142\\
\multicolumn{3}{|c}{:}&\multicolumn{1}{|c|}{:}&\multicolumn{1}{c|}{:}\\
\hline
\end{tabular}
}
\\\\
\multicolumn{2}{c}{
\renewcommand{\arraystretch}{0.9}
\tabcolsep=1.0pt
\begin{tabular}{|lclclcl|r|r|}
\hline
\multicolumn{7}{|c|}{labels for $\cluster_5$}&$\npkx$&$\npxk$\\
\hline
venomous=T&$\land$&legs=4&&&&&0.943&0.239\\
eggs=F&$\land$&milk=F&&&&&1.000&0.199\\
milk=F&$\land$&toothed=T&$\land$&fin=F&&&1.000&0.799\\
hair=F&$\land$&toothed=T&$\land$&fin=F&&&0.935&0.799\\
milk=F&$\land$&toothed=T&$\land$&breathes=T&&&1.000&0.719\\
milk=F&$\land$&breathes=T&$\land$&legs=4&&&1.000&0.539\\
feathers=F&$\land$&milk=F&$\land$&backbone=T&$\land$&fin=F&1.000&0.899\\
hair=F&$\land$&feathers=F&$\land$&backbone=T&$\land$&fin=F&0.931&0.899\\
\multicolumn{7}{|c}{:}&\multicolumn{1}{|c|}{:}&\multicolumn{1}{c|}{:}\\
\hline
\end{tabular}
}
\\\\
\multicolumn{2}{c}{
\renewcommand{\arraystretch}{0.9}
\tabcolsep=1.0pt
\begin{tabular}{|lclcl|r|r|}
\hline
\multicolumn{5}{|c|}{labels for $\cluster_6$}&$\npkx$&$\npxk$\\
\hline
backbone=F&$\land$&breathes=T&&&0.916&1.000\\
predator=F&$\land$&backbone=F&&&0.978&0.899\\
breathes=T&$\land$&legs=6&&&1.000&0.800\\
aquatic=F&$\land$&legs=6&&&0.965&0.800\\
predator=F&$\land$&legs=6&&&1.000&0.720\\
airborne=T&$\land$&backbone=F&&&1.000&0.600\\
feathers=F& $\land$&eggs=T&$\land$&airborne=T&1.000&0.600\\
feathers=F&$\land$&airborne=T&$\land$&toothed=F&1.000&0.600\\
\multicolumn{5}{|c}{:}&\multicolumn{1}{|c|}{:}&\multicolumn{1}{c|}{:}\\
\hline
\end{tabular}
}
\end{tabular}
}
\end{table}

\subsection{Iris dataset}
\label{sec:experiment:iris}

\begin{figure}[t]
\centerline{
\begin{tabular}{cp{10pt}c}
\includegraphics[width=0.35\columnwidth]{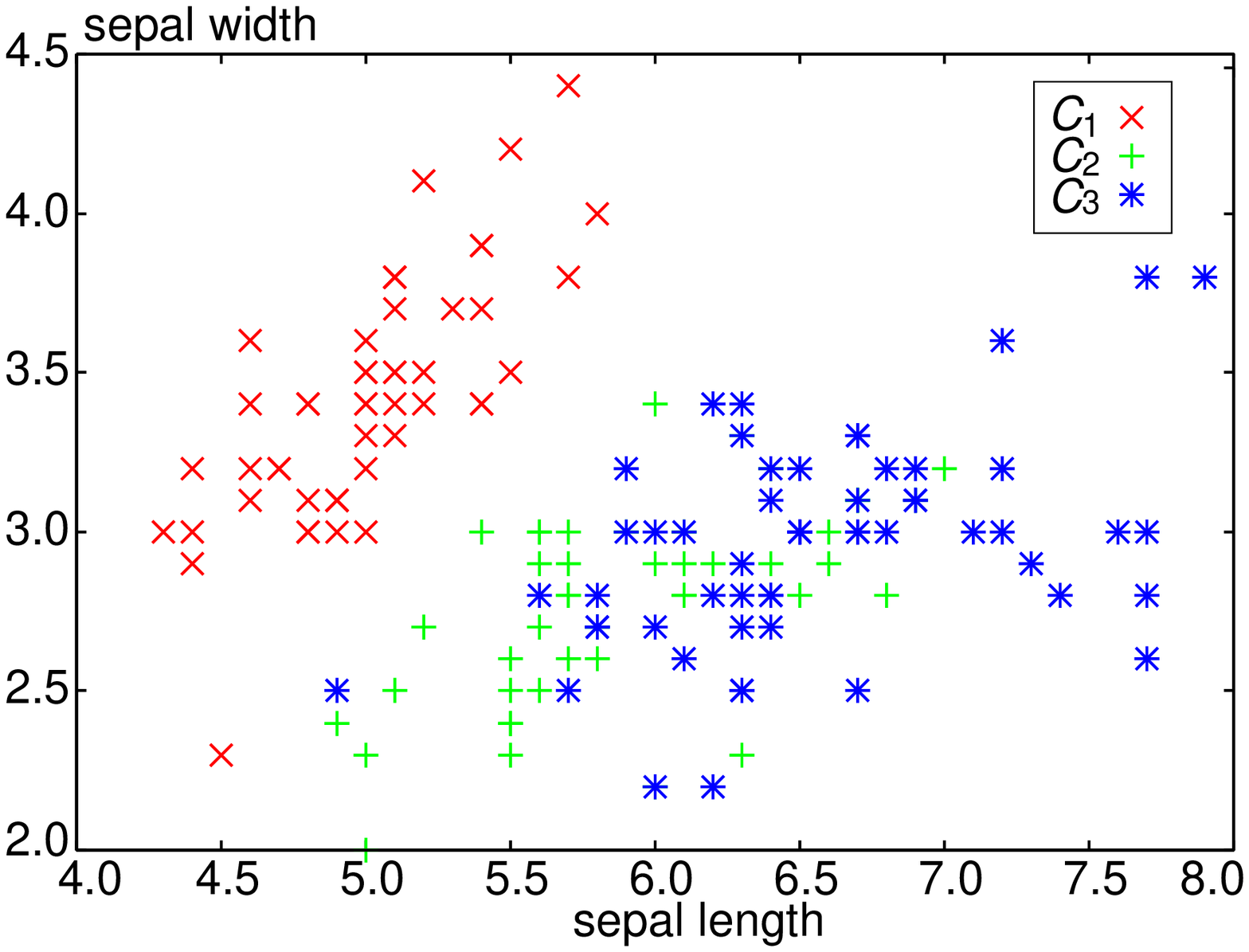}
&&
\includegraphics[width=0.35\columnwidth]{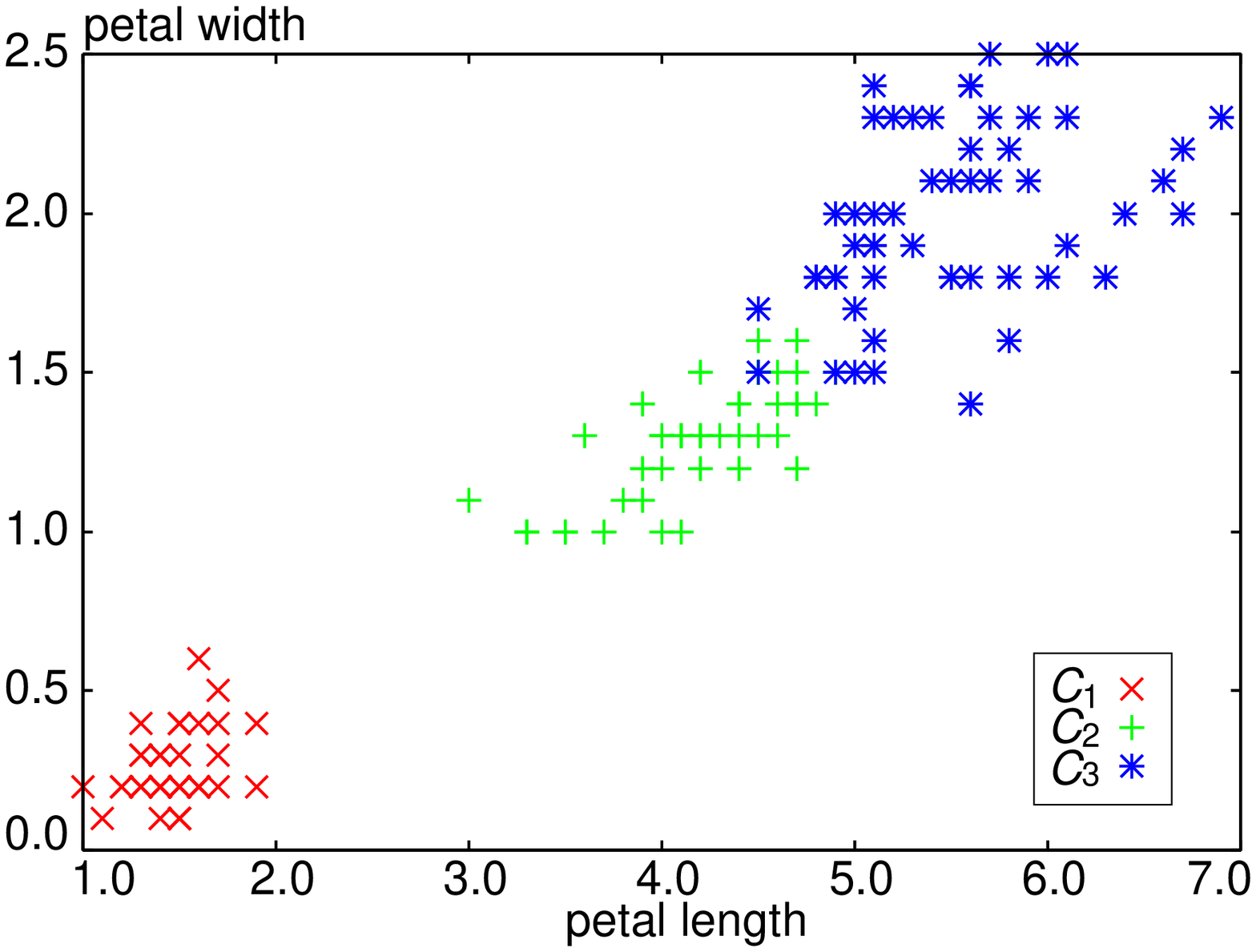}
\end{tabular}
}
\caption{
Scattered plots of the iris dataset (right) for sepal-length
vs.\ sepal-width and (left) for petal-length vs.\ petal-width.
}
\label{fig:iris}
\end{figure}

As a typical continuous dataset, we picked up the iris dataset, in which
there are four attributes:\ petal width, petal length, sepal width and
sepal length.  Each of 150 cases in the dataset originally belongs to one of
three classes:\ Setosa, Versicolour and Virginica.  The confusion matrix and
the obtained labels are shown in Table~\ref{tab:iris}.
We used the thresholds $r=0.9$ and $\slocal=K/|\dataset|$.
A candidate set $\cprobs$ of cumulative probabilities
(introduced in Section~\ref{sec:proposal:cont}) is $\{0.2, 0.4, 0.6, 0.8\}$.
The scattered plots in Fig.~\ref{fig:iris} tell us that
the obtained characteristic labels adaptively capture
the dense part of cluster $\cluster_1$.
Also it should be noted that, in the proposed method, the Euclidean distance
from the center of the cluster is translated into a cumulative probability
under a Gaussian distribution.

\begin{table}[t]
\caption{
(top) The confusion matrix in clustering the iris dataset, and
(bottom) the obtained labels.
}
\centerline{
\begin{tabular}{c}
{
\tabcolsep=3.5pt
\begin{tabular}{|r|rrr|}
\hline
&\multicolumn{3}{|c|}{clusters}\\
original classes&$\cluster_1$&$\cluster_2$&$\cluster_3$\\
\hline
Setosa  &50& 0& 0\\
Versicolour& 0&45& 5\\
Virginica  & 0& 0&50\\
\hline
\end{tabular}
}
\\
\\
{\footnotesize
\renewcommand{\arraystretch}{0.85}
\tabcolsep=3.0pt
\begin{tabular}{|rcl|r|r|}
\hline
\multicolumn{3}{|c|}{labels for $\cluster_1$}&$\npkx$&$\npxk$\\
\hline
$0.06<\mbox{petal-w}\le 0.43$&&&0.999&0.800\\
$1.2<\mbox{petal-l}\le 1.7$&&&1.000&0.799\\
$3.3<\mbox{sepal-w}\le 3.5$&&&0.953&0.199\\
$4.3<\mbox{sepal-l}\le 5.6$ &$\land$&$3.0<\mbox{sepal-w}\le 3.8$&0.978&0.480\\
$4.9<\mbox{sepal-l}\le 5.1$&$\land$&$2.8< \mbox{sepal-w}\le 4.0$&0.926&0.159\\
\hline
\multicolumn{3}{}{}\\
\hline
\multicolumn{3}{|c|}{labels for $\cluster_2$}&$\npkx$&$\npxk$\\
\hline
$4.0<\mbox{petal-l}\le 4.4$&&&0.931&0.799\\
$5.5<\mbox{sepal-l}\le 6.4$&$\land$&$0.99<\mbox{petal-w}\le 1.6$&0.979&0.479\\
$5.8<\mbox{sepal-l}\le 6.0$&$\land$&$2.6<\mbox{sepal-w}\le 2.9$& 0.964&0.160\\ 
\hline
\multicolumn{3}{}{}\\
\hline
\multicolumn{3}{|c|}{labels for $\cluster_3$}&$\npkx$&$\npxk$\\
\hline
$4.5<\mbox{petal-l}\le 6.4$&&&0.985&0.800\\
$6.4<\mbox{sepal-l}\le 6.7$&&&0.904&0.800\\
$1.7<\mbox{petal-w}\le 2.2$&&&0.942&0.400\\
$2.8<\mbox{sepal-w}\le 3.1$&$\land$&$1.6<\mbox{petal-w}\le 2.4$&0.918&0.480\\
$2.8<\mbox{sepal-w}\le 4.0$&$\land$&$1.6<\mbox{petal-w}\le 2.4$&0.947&0.446\\
\hline
\end{tabular}
}
\end{tabular}
}
\label{tab:iris}
\end{table}

\subsection{20 newsgroups dataset}
\label{sec:experiment:20news}

The 20 newsgroups dataset is originally a collection of approximately
20,000 articles from 20 different newsgroups.
A preprocessed dataset available from
{\tt http://\linebreak[3]people.\linebreak[3]csail.\linebreak[3]mit.\linebreak[3]edu/jrennie/\linebreak[3]20Newsgroups/} is used,
and the articles from three newsgroups:
{\tt comp.\linebreak[3]sys.\linebreak[3]ibm.\linebreak[3]pc.\linebreak[3]hardware}, {\tt rec.\linebreak[3]sport.\linebreak[3]hockey} and
{\tt soc.\linebreak[3]religion.\linebreak[3]christian}.
We made further preprocessing: stemming by
the Porter's algorithm~\cite{Manning08},
removing infrequent words ($\le 200$ occurrences), removing short articles
($\le 10$ words) and removing the attributes taking only one value.
The dataset was finally converted into 2,799 bag-of-words boolean
vectors whose dimension is 2,016.
In labeling by the proposed method, we did not use the conjuncts
of the form $\quoted{w=\false}$
(or $\quoted{w=\mathrm{F}}$) which means the absence of word
$w$ in the article.  The thresholds $r$ and
$\slocal$ were respectively configured as $0.9$ and
$10\times K/|\dataset|$.\footnote{
$\slocal$ was configured as $10\times K/|\dataset|$ because
the 20 newsgroup dataset is 10 times (or more) larger than the zoo
and the iris dataset.
}
Furthermore, we applied the greedy
pruning described at the last of Section~\ref{sec:proposal:search}.

The results are shown in Table~\ref{tab:20news}.
From the obtained characteristic labels for $\cluster_1$,
it is seen that the article containing words such as
``hockei'' (``hockey'';
the suffix should have been replaced by the stemmer) and ``nhl'' (``NHL'';
the National Hockey League)
are likely to belong to $\cluster_1$.
There are also the names of a hockey team and its home city
(i.e.\ Pittsburgh Penguins).
So we can guess from this information that
$\cluster_1$ is a cluster of articles related to hockey.
Similarly, it is easy to see that $\cluster_2$ is a cluster of
articles related to computer hardware,\footnote{
As shown in the confusion matrix in Table~\ref{tab:20news},
$\cluster_2$ contains the articles from {\tt soc.\linebreak[3]religion.\linebreak[3]christian},
but the characteristic labels related to religion did not appear.
This would be because the articles from {\tt soc.\linebreak[3]religion.\linebreak[3]christian}
mainly use non-technical terms, which are less likely to form
characteristic labels.
}
from the words such as
``mb'' (``megabytes'' or ``motherboard''), ``disk'' and
``motherboard.''
$\cluster_3$ would be understood as a cluster
that contains the articles talking about religious matters.
Although there are many attributes in this dataset, our search algorithm
is feasible,\footnote{
It took 404 seconds on a PC with Core i7 2.66GHz to get all characteristic
labels for all clusters.
Currently the search algorithm is implemented in the Ruby script language.
}
thanks to the pruning based on the minimality and the optimized setting
described above.

\begin{table}[t]
\caption{
The confusion matrix,
and the characteristic labels for the clusters $\cluster_1$,
$\cluster_2$ and $\cluster_3$ in the 20 newsgroups dataset.
}
\label{tab:20news}
{
\tabcolsep=0pt
\begin{center}
{\footnotesize
\tabcolsep=3pt
\begin{tabular}{|l|rrr|}
\hline
&\multicolumn{3}{|c|}{clusters}\\
\multicolumn{1}{|c|}{original classes}&
\multicolumn{1}{c}{$\cluster_1$}&
\multicolumn{1}{c}{$\cluster_2$}&
\multicolumn{1}{c|}{$\cluster_3$}\\
\hline
{\tt comp.sys.ibm.pc.hardware}&  0&907&  7\\
{\tt rec.sport.hockey}        &899& 28& 10\\
{\tt soc.religion.christian}  &  2&371&575\\
\hline
\end{tabular}
}
\par\vspace{5pt}\par
\begin{tabular}{ccc}
{
\footnotesize
\renewcommand{\arraystretch}{0.9}
\tabcolsep=3pt
\begin{tabular}[t]{|l|r|r|}
\hline
\multicolumn{1}{|c|}{labels for $\cluster_1$}&$\npkx$&$\npxk$\\
\hline 
game=T      &0.930&0.552\\
team=T      &0.976&0.487\\
hockei=T    &0.963&0.381\\
player=T    &0.959&0.319\\
playoff=T   &0.969&0.278\\
season=T    &0.964&0.248\\
nhl=T       &0.989&0.217\\
cup=T       &0.927&0.200\\
score=T     &0.936&0.198\\
leagu=T     &0.968&0.174\\
wing=T      &0.905&0.159\\
pittsburgh=T&0.956&0.149\\
toronto=T   &0.922&0.145\\
leaf=T      &0.968&0.137\\
detroit=T   &0.983&0.135\\
bruin=T     &0.990&0.134\\
penguin=T   &0.982&0.134\\
\multicolumn{1}{|c|}{:}&\multicolumn{1}{|c|}{:}&\multicolumn{1}{|c|}{:}\\
knock=T     &0.916&0.013\\
year=T   $\land$ plai=T&0.919&0.147\\
ca=T     $\land$ plai=T&0.932&0.125\\
articl=T $\land$ fan=T &0.902&0.121\\
plai=T   $\land$ win=T &0.985&0.115\\
\multicolumn{1}{|c|}{:}&\multicolumn{1}{|c|}{:}&\multicolumn{1}{|c|}{:}\\
\hline
\end{tabular}
}
&
{
\footnotesize
\renewcommand{\arraystretch}{0.9}
\tabcolsep=3pt
\begin{tabular}[t]{|l|r|r|}
\hline
\multicolumn{1}{|c|}{labels for $\cluster_2$}&$\npkx$&$\npxk$\\
\hline
card=T       &0.959&0.217\\
pc=T         &0.972&0.167\\
mb=T         &0.993&0.132\\
bu=T         &0.929&0.132\\
disk=T       &0.969&0.124\\
window=T     &0.958&0.124\\
instal=T     &0.926&0.106\\
driver=T     &0.903&0.094\\
motherboard=T&0.990&0.092\\
ibm=T        &0.966&0.092\\
\multicolumn{1}{|c|}{:}&\multicolumn{1}{|c|}{:}&\multicolumn{1}{|c|}{:}\\
batteri=T    &0.993&0.011\\
drive=T $\land$ work=T&0.903&0.053\\
drive=T $\land$ system=T&0.971&0.049\\
\multicolumn{1}{|c|}{:}&\multicolumn{1}{|c|}{:}&\multicolumn{1}{|c|}{:}\\
\hline
\end{tabular}
}
&
{
\footnotesize
\renewcommand{\arraystretch}{0.9}
\tabcolsep=3pt
\begin{tabular}[t]{|l|r|r|}
\hline
\multicolumn{1}{|c|}{labels for $\cluster_3$}&$\npkx$&$\npxk$\\
\hline
divin=T    &0.950&0.099\\
fals=T     &0.904&0.097\\
condemn=T  &0.904&0.096\\
reveal=T   &0.931&0.094\\
societi=T  &0.921&0.081\\
kingdom=T  &0.920&0.068\\
guilti=T   &0.963&0.049\\
innoc=T    &0.932&0.049\\
israel=T   &0.959&0.043\\
social=T   &0.942&0.030\\
diseas=T   &0.909&0.018\\
islam=T    &0.909&0.018\\
jehovah=T  &0.989&0.015\\
\multicolumn{1}{|c|}{:}&\multicolumn{1}{|c|}{:}&\multicolumn{1}{|c|}{:}\\
explor=T   &0.986&0.012\\
christian=T $\land$ god=T&0.954&0.437\\
peopl=T $\land$ god=T&0.928&0.435\\
\multicolumn{1}{|c|}{:}&\multicolumn{1}{|c|}{:}&\multicolumn{1}{|c|}{:}\\
\hline
\end{tabular}
}
\end{tabular}
\end{center}
}
\end{table}

\subsection{Flags dataset}
\label{sec:experiment:flags}

The flags dataset contains the details of 194 national flags, originally
described by 30 attributes.  In this experiment, we focused on the clusters
of national flags grouped on their visual aspects, and hence non-visual
attributes (landmass, zone, area, population, language and religion) were
removed in advance.  As is written above, since the class
information is not given in this dataset, we first estimated the number of
clusters as $\hat{K}$ by the Cheeseman-Stutz score~\cite{Cheeseman95},
a Bayesian model selection criterion adopted in AutoClass, and then
starting from $\hat{K}$, we explored a plausible number of clusters
by observing the characteristic labels.
Another point in this dataset is that discrete attributes and continuous
attributes are mixed.  That is, all of eight integer attributes
(e.g.\ the number of circles in the flag) were treated
as continuous attributes.  We used $r=0.75$ and $\slocal=K/|\dataset|$
as the thresholds for $\pkx$ and $\pxk$, respectively, where
$\dataset$ is the dataset and $K$ is the number of clusters.
Also we conducted the greedy pruning.

\begin{figure}[t]
\centerline{
\includegraphics[bb=50 66 280 159,clip,scale=1.1]{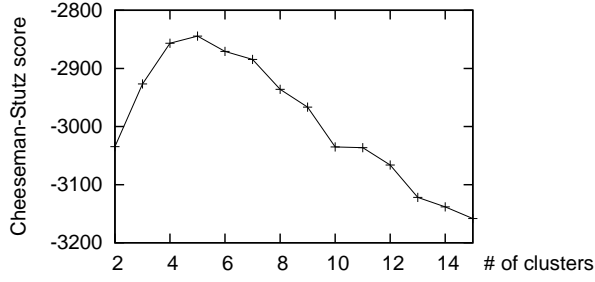}
}
\caption{The Cheeseman-Stutz scores with various numbers of clusters.}
\label{fig:score}
\end{figure}

Fig.~\ref{fig:score} shows the curve of the Cheeseman-Stutz score with various
numbers of clusters, and we have $\hat{K}=5$ as a peak of this curve.
We further continued to compute characteristic labels with the number $K$
of clusters being around $\hat{K}$, and found that readable characteristic
labels are obtained when $K=6$.
Table~\ref{tab:flags} presents these labels.\footnote{
Since each continuous attribute $A_j$ is originally an integer attribute,
a proposition ``$\alpha < A_j \le \beta$'' (assume here that
$\alpha$ and $\beta$ are not integers, for simplicity) was translated back into
``$A_j=\lceil\alpha\rceil,\lceil\alpha\rceil+1,\ldots,\lfloor\beta\rfloor$''
in Table~\ref{tab:flags}.  Non-minimal labels produced by this
translation were then removed.
}
The shortest characteristic label for the cluster $\cluster_1$ 
says that the national flags in $\cluster_1$ (and none in the other clusters)
have one saltire (diagonal cross).  A typical example of such flags is the
Union Jack, and actually many flags in $\cluster_1$ have one quartered section
(i.e.\ \#quarters=1) for the Union Jack.  Similarly, the clusters $\cluster_2$ and
$\cluster_3$ contain the flags with vertical bars and with circles,
respectively.  The label (\#saltires=0 $\wedge$ \#quarters=1) for $\cluster_6$
distinguishes $\cluster_1$ and $\cluster_6$,
and similarly the labels (\#crosses=1 $\wedge$ \#saltires=0) and
(\#crosses=1 $\wedge$ \#quarters=0) for $\cluster_4$ jointly work for
distinguishing $\cluster_4$ from $\cluster_1$ and $\cluster_6$,
where \#crosses indicates the number of upright crosses.
Indeed, $\cluster_6$ contains the flag of the United States,
and $\cluster_4$ contains the flags of
several Scandinavian countries (note that the Union Jack also contains
upright crosses).
From the labels for $\cluster_5$, one may see
that $\cluster_5$ is a cluster of miscellaneous flags.
On the other hand, when the number $K$ of clusters is set at $\hat{K}=5$,
the clusters $\cluster_2$ and $\cluster_3$ are merged into one cluster,
whose characteristic labels are not so intuitive as in Table~\ref{tab:flags}.
These results imply that a plausible number of clusters can be
determined by interactively consulting characteristic labels,
with a help from model selection techniques, and clearly exemplify
how the feedbacks from the interpretation/evaluation step contribute
in knowledge discovery.

\section{Related work}
\label{sec:related}

As mentioned above, there have been only a few labeling approaches.
LabelSOM~\cite{Rauber99} is a labeling method for self-organizing maps,
and Mei et al.'s automatic labeling method for unigram topic models~\cite{Mei07}
uses a heuristic score based on pointwise mutual information.  As described
in Section~\ref{sec:proposal:cpl}, different relevance measures are used
by Popescul and Ungar~\cite{Popescul00} and by Lamirel et al.~\cite{Lamirel08}
for automatic labeling of document clusters.  In these labeling methods, the length
of possible labels seems to be limited in advance, and thus no pruning mechanism,
like the one described in Section~\ref{sec:proposal:search}, is given.

\begin{table}[t]
\caption{
The characteristic labels for $\cluster_1$, \ldots, $\cluster_6$ in the flags dataset.
}
\label{tab:flags}
{
\tabcolsep=0pt
\begin{center}
\begin{tabular}{cc}
{\footnotesize
\renewcommand{\arraystretch}{0.9}
\tabcolsep=3pt
\begin{tabular}{|l|r|r|}
\hline
\multicolumn{1}{|c|}{labels for $\cluster_1$}&$\npkx$&$\npxk$\\
\hline 
\#saltires=1&1.000&0.900\\
topleft=white $\wedge$ \#quarters=1&0.817&0.622\\
stripes=0,1,2 $\wedge$ \#quarters=1&0.827&0.540\\
botright=blue $\wedge$ \#quarters=1&0.819&0.505\\
green=T $\wedge$ \#crosses=1&0.906&0.467\\
gold=T $\wedge$ \#crosses=1&0.763&0.467\\
mainhue=blue $\wedge$ \#quarters=1&0.810&0.467\\
\#crosses=1 $\wedge$ \#quarters=1&0.751&0.420\\
\multicolumn{1}{|c|}{:}&\multicolumn{1}{|c|}{:}&\multicolumn{1}{|c|}{:}\\
\hline
\multicolumn{3}{c}{}\\
\hline
\multicolumn{1}{|c|}{labels for $\cluster_2$}&$\npkx$&$\npxk$\\
\hline 
\#bars=1,2,3,4&0.782&0.800\\
\hline
\multicolumn{3}{c}{}\\
\hline
\multicolumn{1}{|c|}{labels for $\cluster_3$}&$\npkx$&$\npxk$\\
\hline 
\#circles=1,2 $\wedge$ \#crosses=0&0.781&0.540\\
\#circles=1,2 $\wedge$ \#quarters=0&0.781&0.540\\
black=T $\wedge$ \#circles=1&0.766&0.225\\
blue=F $\wedge$ \#circles=1&0.765&0.200\\
botright=green $\wedge$ \#circles=1,2&0.781&0.181\\
topleft=orange $\wedge$ \#saltires=0&0.999&0.135\\
topleft=orange $\wedge$ \#crosses=0&0.970&0.135\\
mainhue=orange $\wedge$ \#crosses=0&0.970&0.135\\
\multicolumn{1}{|c|}{:}&\multicolumn{1}{|c|}{:}&\multicolumn{1}{|c|}{:}\\
\hline
\end{tabular}
}
&
{\footnotesize
\renewcommand{\arraystretch}{0.9}
\tabcolsep=3pt
\begin{tabular}{|l|r|r|}
\hline
\multicolumn{1}{|c|}{labels for $\cluster_4$}&$\npkx$&$\npxk$\\
\hline 
\#crosses=1 $\wedge$ \#saltires=0&0.810&\multicolumn{1}{|l|}{\hq 0.81003}\\
\#crosses=1 $\wedge$ \#quarters=0&0.829&\multicolumn{1}{|l|}{\hq 0.81002}\\
\#crosses=1 $\wedge$ \#sunstars=0&0.751&\multicolumn{1}{|l|}{\hq 0.720}\\
\#circles=0 $\wedge$ \#crosses=1&0.768&\multicolumn{1}{|l|}{\hq 0.640}\\
green=F $\wedge$ \#crosses=1&0.757&\multicolumn{1}{|l|}{\hq 0.500}\\
\#colors=2,3 $\wedge$ \#crosses=1&0.759&\multicolumn{1}{|l|}{\hq 0.490}\\
gold=F $\wedge$ \#crosses=1&0.754&\multicolumn{1}{|l|}{\hq 0.356}\\
\hline
\multicolumn{3}{c}{}\\
\hline
\multicolumn{1}{|c|}{labels for $\cluster_5$}&$\npkx$&$\npxk$\\
\hline 
\#bars=0&0.803&0.900\\
\#circles=0&0.752&0.900\\
\#crosses=0&0.755&0.600\\
\#quarters=0&0.752&0.400\\
triangle=T&0.889&0.240\\
botright=black&0.888&0.080\\
mainhue=black&0.799&0.040\\
\multicolumn{1}{|c|}{:}&\multicolumn{1}{|c|}{:}&\multicolumn{1}{|c|}{:}\\
\hline
\multicolumn{3}{c}{}\\
\hline
\multicolumn{1}{|c|}{labels for $\cluster_6$}&$\npkx$&$\npxk$\\
\hline 
\#saltires=0 $\wedge$ \#quarters=1&0.960&0.360\\
topleft=blue $\wedge$ \#quarters=1&0.875&0.320\\
\hline 
\end{tabular}
}
\end{tabular}
\end{center}
}
\end{table}

CLIQUE~\cite{Agrawal98} is a novel hyper-rectangular
clustering method that additionally gives comprehensible descriptions
of the obtained clusters.  The description of each cluster is
a DNF formula of the ranges of continuous attributes such as
$((30\le\mathrm{age}<50)\land(4\le\mathrm{salary}<8))
\lor((40\le\mathrm{age}<60)\land(2\le\mathrm{salary}<6))$.
Although CLIQUE has a similar motivation to ours, it is mainly designed
for the dataset with continuous attributes.  According to the original
description (``Remarks'' in Section 2.2 of \cite{Agrawal98}), if we use
discrete attributes, all instances in a cluster must take the same value
for each discrete attribute in a selected subspace.
In the proposed method, contrastingly, we do not have such a restriction,
and as seen in Section~\ref{sec:experiment:flags},
we can make use of advanced statistical
techniques such as ones for model selection in the clustering step.
The latter point also contrasts the proposed method with conceptual clustering
methods such as COBWEB~\cite{Fisher87}.

In the research on expert systems, it has been a problem to explain the
expert system's conclusion to human users.  Wolverton~\cite{Wolverton95} proposed
the use of satisficing conclusion-substantiating (SCS) explanations
to explain an expert system's conclusion.  Given a system's conclusion
$c$ and a threshold $\rho$, the SCS explanation $\ve$ is the shortest
sequence of facts such that $p(c\mid\ve)>\rho$
(or if no such sequence of facts, $\ve=\mathrm{argmax}_{\fve'\;}p(c\mid\ve')$).
Our search algorithm would contribute in efficient finding of SCS explanations.

Traditional rule induction methods such as C4.5 and RIPPER~\cite{Witten05}
can also be applied to find comprehensible cluster descriptions.
However, Hotho et al.\ reported that these methods tend to
produce too many rules to manage for human~\cite{Hotho03}.
One possible reason is that C4.5 and RIPPER have
a representational limitation that the premises in the obtained rules are
always exclusive and need to be understood fragmentarily.
In the proposed method, on the other hand, each characteristic label is
independently interpretable.
Another possibility is that C4.5 and RIPPER tried to find the exact boundaries
among clusters, by their design.  In labeling, however, we do not always
have to find such exact boundaries.
Furthermore, traditional rule induction methods often suffer from a so-called
rare-class problem~\cite{Weiss04} when we have imbalanced or
many clusters (if there are many clusters, each cluster is relatively rare).
For example, small groups of instances (small disjuncts) in a rare class
are often missed.  Actually, in the zoo dataset,
C4.5/RIPPER only generated the rules for the cluster $\cluster_3$ (``birds''):
$\quoted{\mbox{feathers=True}}\Rightarrow\cluster_3$,
and
$\quoted{\mbox{feathers=False}}\Rightarrow\neg\cluster_3$, and
the rest of the antecedent patterns we found (Table~\ref{tab:zoo:1},
bottom-right)
were ignored.
This is presumably because most of the instances have been covered
by the simple rules above in the rule construction process of C4.5/RIPPER.
It is reported that a classifier based on emerging patterns
works well for the rare-class problem~\cite{Ramamohanarao05}.

Recently it is proposed in \cite{KraljNovak09} to unify three similar
data mining tasks, contrast set mining, emerging pattern mining and
subgroup discovery, under the name of
{\em supervised descriptive rule discovery}.
Our labeling method can be seen as a model-based approach in this
framework, which focuses on interpretation/evaluation of probabilistic
clusters.  In a broader context, for knowledge discovery under an unsupervised
setting, a sequential run of clustering and {\em discriminative} labeling
would be a promising alternative to {\em frequent} pattern mining.
Besides, also recently, Zimmermann and De Raedt introduced a general
data mining task called {\em cluster-grouping}~\cite{Zimmermann09}, and
a branch-and-bound algorithm, named CG, for this task.
CG efficiently finds characteristic patterns (labels, in our case)
following a guide from a convex relevance score such as $\chi^2$, information gain
(used in ID3), WRAcc (Section~\ref{sec:proposal:cpl:relv})
and category utility (used in COBWEB).  Although this algorithm is powerful,
it could not be directly applied to our labeling problem, since
the membership probability $\pkx$ seems not convex.

In the context of probabilistic modeling, the proposed method with mixture
models could be extended for evidence-based sensitivity analysis
(e.g.\ \cite{Jensen96}) or explanatory analysis (e.g.\ \cite{Chajewska97})
of Bayesian networks, in which the membership probability $\pkx$ is generalized
as $p(q\mid\ve)$, where $q$ is an instantiation of a query variable and $\ve$ is
an instantiation of (a part of) evidence variables, and thus we search for
a minimal combination $\ve$ of evidences which is highly influential to the
observation $q$.  To the best of our knowledge, the most recent and closest
work is Yuan et al.'s general framework for {\em most relevant explanation}
(MRE)~\cite{Yuan08,Yuan09}.  Their MRE framework adopts a relevance score called
generalized Bayes factor (GBF), defined as $\gbf_k(\vx)=p(k\mid\vx)/p(k\mid\neg\vx)$
in our labeling problem.  The MRE framework looks attractive, but seems
unfit to our case for a couple of reasons.
First,
for the $k$-th cluster,
a ranking over the propositional labels $\vx$ by $\gbf_k(\vx)$ is different
from the one used in the clustering step (i.e.\ by $\pkx$).
Second,
$\gbf_k(\vx)=\frac{1-p(\fvx)}{1-p(\fvx\mid k)}\cdot\frac{p(\fvx\mid k)}{p(\fvx)}$
can be numerically unstable when $p(\vx\mid k)\approx 1$.   For instance,
we cannot order the labels $\vx$ such that $p(\vx\mid k)=1$, which in fact
appear in one of our experiments (i.e.\ Table~\ref{tab:zoo:1}).  Third,
the MRE framework only provides an MCMC-based approximate method or an
exact (exhaustive) method without safe pruning (like the one based
on global/local support and minimality in the proposed method) for finding
relevant $\vx$.
Lastly, the MRE papers do not describe how to handle continuous attributes.

Handling continuous attributes is an important issue in CAR
(class association rule) mining.
For example, Washio et al.~\cite{Washio07} proposed a CAR mining method
that discretizes the continuous space on the fly with hyper-rectangular
clustering.  The difference from our labeling method is that we are given
probabilistic clusters from beginning and thus we effectively limit
propositions to the ones of the form
$\quoted{\alpha< A_j\le \beta}$, where $\alpha$ and $\beta$ are
symmetric w.r.t.\ the mean in the cluster.
Besides, as in usual CAR mining, Washio et al.'s method searches for the
antecedent patterns $\vx$ based on the local support $\pxk$.

Section~\ref{sec:proposal:overview} described that the EM algorithm
is adopted for clustering.  We can also use the $K$-means algorithm instead,
since $K$-means can be seen as an instance of a parameter estimation framework
often called {\em Viterbi training},\footnote{
Viterbi training has also been called hard EM, Viterbi EM, classification EM,
sparse EM, and so on.
}
tailored for a Gaussian mixture model with equal class probabilities
and a common covariance matrix of the form
$\sigma^2I$~\cite{Celeux92}.  Once the model parameters have been
estimated, our labeling method is applicable as written in this paper.
Similarly to the case in Section~\ref{sec:experiment:iris}, when combined with
$K$-means, the Euclidean distance from the centroid (the mean in the cluster)
is translated into a cumulative probability under a Gaussian distribution.

\section{Conclusion and future work}
\label{sec:conclusion}

In this paper, we proposed a new labeling method that associates propositional
labels (conjunctions of attribute-value pairs) with the clusters obtained by
mixture models, to help us interpret or evaluate the clusters.
As shown in the experimental results, the proposed
method finds a set of intuitive descriptive labels that characterize well
or ``verbalize'' the clusters.
The proposed method is fully applicable to various datasets
including continuous attributes and missing values, and can be
a new, in-depth and consistent tool for cluster interpretation/evaluation.
Besides, the experimental results also show that the feedbacks from
the interpretation/evaluation step can play an important role for
achieving a reasonable clustering result.
In future work, we would like to extend the proposed method to use
disjunctive formulas or a richer representation.
For example, we may merge two similar characteristic labels
(milk=T $\land$ legs=4) and (hair=T $\land$ legs=4) into
((milk=T $\lor$ hair=T) $\land$ legs=4) to gain a higher local support.
In a purely logical sense, our labeling
algorithm can be formulated under the setting of inductive logic programming
(ILP) with a simple refinement operator.
As in ILP, the use of background knowledge such as taxonomy seems helpful
for having more comprehensible descriptions.

\section*{Acknowledgments.}
The authors would like to thank Toshihiro Kamishima for his helpful
comments on related work.
This work is supported in part by Grant-in-Aid for Scientific Research
(No.~20240016)
from Ministry of Education, Culture, Sports, Science and Technology of Japan.

\bibliographystyle{splncs}
\bibliography{draft}

\end{document}